%% file: main.tex
\documentclass[conference]{IEEEtran}
\usepackage{amsmath,amssymb,amsfonts}

\usepackage{caption}
\usepackage{subcaption}

\usepackage{algorithmic}
\usepackage{graphicx}
\usepackage{textcomp}
\usepackage{xcolor}
\usepackage[hidelinks]{hyperref} 
\usepackage{todonotes}
\usepackage[
backend=biber,
style=ieee,
maxnames=5,
minnames=1
]{biblatex}
\usepackage{braket}
\usepackage[capitalize,noabbrev]{cleveref}
\usepackage{multirow}
\usepackage{tabularx}
\usepackage[nohyperlinks,nolist]{acronym}
\usepackage{tikz}
\usetikzlibrary{quantikz}

\newcommand\copyrighttext{%
  \footnotesize \textcopyright 2025 IEEE. Personal use of this material is permitted.
  Permission from IEEE must be obtained for all other uses, in any current or future
  media, including reprinting/republishing this material for advertising or promotional purposes, creating new collective works, for resale or redistribution to servers or lists, or reuse of any copyrighted component of this work in other works. 
  }
\newcommand\copyrightnotice{%
\begin{tikzpicture}[remember picture,overlay]
\node[anchor=south,yshift=10pt] at (current page.south) {\fbox{\parbox{\dimexpr\textwidth-\fboxsep-\fboxrule\relax}{\copyrighttext}}};
\end{tikzpicture}%
}

\def\BibTeX{{\rm B\kern-.05em{\sc i\kern-.025em b}\kern-.08em
    T\kern-.1667em\lower.7ex\hbox{E}\kern-.125emX}}
\begin{document}

\title{Old Rules in a New Game: Mapping Uncertainty Quantification to Quantum Machine Learning}

\author{\IEEEauthorblockN{Maximilian Wendlinger\IEEEauthorrefmark{1}, Kilian Tscharke\IEEEauthorrefmark{1}, Pascal Debus\IEEEauthorrefmark{1}}
\IEEEauthorblockA{\IEEEauthorrefmark{1}\textit{Quantum Security Technologies} \\
\textit{Fraunhofer Institute for Applied and Integrated Security}\\
Garching near Munich, Germany}
\IEEEauthorblockA{\{maximilian.wendlinger, kilian.tscharke, pascal.debus\}@aisec.fraunhofer.de}
}

\maketitle
\copyrightnotice

\begin{abstract}
One of the key obstacles in traditional deep learning is the reduction in model transparency caused by increasingly intricate model functions, which can lead to problems such as overfitting and excessive confidence in predictions. With the advent of quantum machine learning offering possible advances in computational power and latent space complexity, we notice the same opaque behavior. Despite significant research in classical contexts, there has been little advancement in addressing the black-box nature of quantum machine learning. Consequently, we approach this gap by building upon existing work in classical uncertainty quantification and initial explorations in quantum Bayesian modeling to theoretically develop and empirically evaluate techniques to map classical uncertainty quantification methods to the quantum machine learning domain. Our findings emphasize the necessity of leveraging classical insights into uncertainty quantification to include uncertainty awareness in the process of designing new quantum machine learning models.
\end{abstract}


\input{01-intro}
\input{02-related-work}
\input{03-theoretical-background}
\input{04-mapping-ml-qml}
\input{05-experimental-setup}
\input{06-empirical-evaluations}
\input{10-summary-outlook}

\printbibliography
\clearpage

\input{99-appendix}

\begin{acronym}
    \acro{UQ}[UQ]{Uncertainty Quantification}
    \acro{ML}[ML]{Machine Learning}
    \acro{QC}[QC]{Quantum Computing}
    \acro{QML}[QML]{Quantum Machine Learning}
    \acro{DL}[DL]{Deep Learning}
    \acro{NISQ}[NISQ]{Noisy Intermediate-Scale Quantum}
    \acro{QPE}[QPE]{Quantum Phase Estimation}
    \acro{LLMs}[LLMs]{Large Language Models}
    \acro{VI}[VI]{Variational Inference}
    \acro{MC}[MC]{Monte-Carlo}
    \acro{GP}[GP]{Gaussian Process}
    \acro{OOD}[OOD]{out-of-distribution}
    \acro{ELBO}[ELBO]{Evidence Lower Bound}
    \acro{KL}[KL]{Kullback-Leibler}
    \acro{ECE}[ECE]{Expected Calibration Error}
\end{acronym}


\end{document}

%% file: 01-intro.tex
\section{Introduction and Motivation}
\label{sec:intro}

With the availability of large datasets, \ac{DL} models have found their way into our everyday life, with the latest prominent success being \ac{LLMs} solving complex tasks and responding to given questions in ways that are hardly distinguishable from human answers. As models become more complex and the computed model functions less transparent, it is hard to evaluate what underlying reasoning led to a specific model output.

In parallel, the increasing significance of quantum computers -- with their high-dimensional latent Hilbert space and their power to perform complex calculations on superpositions of quantum states -- has opened the path for \ac{ML} models on quantum computers, or \ac{QML}. As a result, the last decade has seen a tremendous increase in research activity in the area of \ac{QML}, both theoretically and practically (physical quantum systems scaling up to over 1000 qubits are currently being engineered \cite{jaygambetta2023}).

Despite this immense success in engineering and computational science, (quantum) \ac{ML} models have to be \emph{trustworthy} to be helpful in practice. As these models become more tightly integrated into applications related to the safety of human lives, it is of the greatest importance to evaluate the decisions made by some given model in the context of the considered input data distribution. Multiple lines of research try to find interpretable characteristics of ML model output to open up this ``black-box'' state of current \ac{ML}. One such approach is called \textit{\acf{UQ}}, which deals with the idea of quantifying the prediction confidence related to \ac{OOD} data or insufficient model complexity.

While the field of \ac{UQ} in classical \ac{ML} has a broad spectrum of research directions and much work is dedicated to finding solutions to the challenges described above, there is hardly any exploration of \ac{UQ} in the domain of \ac{QML}. Motivated by this lack of investigation, we seek to find mappings from techniques regarding \ac{UQ} in classical \ac{ML} to the \ac{QML} context, which -- as described -- currently enjoys an immense research activity, due to its promising advances in many areas, such as speed-ups in tasks of computational complexity \cite{shor1994, harrow2009, liu2021}  and kernel optimizations using large latent spaces \cite{schuld2021a, schuld2019a, rapp2024}.
Thus, the main contributions of this paper are as follows:
\begin{itemize}
    \item We construct a Bayesian \ac{QML} model building upon insights from \acl{VI} in \ac{QML} \cite{nguyen2022} and extending this line of work by a dedicated focus on \ac{UQ}.
    \item We design different quantum dropout methods and their relationship to \ac{MC} dropout in classical \ac{ML}, where no prior work has investigated the effects of any of the chosen methods in the given \ac{MC} dropout frame.
    \item We extend the dropout mechanism to quantum ensembles. While initial works show how to integrate ensemble approaches into \ac{QML} \cite{schuld2017, nguyen2022}, no effort has been made towards the aspects of uncertainty, a gap we fill here.
    \item Lastly, we explore quantum \ac{GP} models using insights from \cite{zhao2019a, rapp2024}, and extend their initial investigations by introducing \ac{UQ} evaluations.
    \item All of the mentioned techniques above are comparatively evaluated regarding ease of implementation, quality of uncertainty estimation, and training/evaluation overhead.
\end{itemize}

%% file: 02-related-work.tex
\section{Related Work}
\label{sec:related-work}

As described, the applications of \ac{UQ} in the field of \emph{classical} \ac{ML} have been thoroughly investigated and are still topic of relevant research. An overview of recent techniques can be found in \cite{galUncertaintyDeepLearning2016, he2024a, nemani2023}. Specific details about classical \ac{UQ} techniques are given in the theoretical background (\cref{sec:theo-back}), along with corresponding sources.
In contrast, the intersection of quantum computing, machine learning, and uncertainty quantification is just beginning to be explored. 

A first approach towards such an exploration is done by \citeauthor{park2023} \cite{park2023}, using quantum conformal prediction to augment given model outputs with ``error bars'' using an additional (held-out) calibration dataset. However, as the basis of their approach is the post-hoc calibration of QML circuits, the model itself is not equipped with the ability to perform predictive inference, which is the focus of this work.

From this predictive uncertainty perspective, \citeauthor{nguyen2022} (\citeyear{nguyen2022}) focus on the idea of combining the fields of \ac{QML} and Bayesian learning to enhance the quantum model's generalization performance and uncertainty estimation. To this end, the authors investigate \emph{BayesByBackprop} and \emph{Deep Ensembles} concerning its applicability in \ac{QML}.
While the authors note the positive effects of the Bayesian learning procedure on the model's capabilities, their work lacks a dedicated evaluation of the uncertainty estimates provided by this technique, a gap we fill in this paper.

Related to our construction of \ac{MC} dropout for \ac{QML}, \citeauthor{kobayashi2022} (\citeyear{kobayashi2022}) and \citeauthor{scala2023} (\citeyear{scala2023}) present the idea of using dropout in quantum circuit architectures. While these works focus on the regularization properties of dropout in \ac{QML}, we leverage similar concepts during inference to approximate the predictive posterior distribution as done in classical \ac{UQ} \cite{gal2016}.

Lastly, as there exist many connections between \ac{QML} and kernel methods, the combination of \ac{GP} and \ac{QML} has attracted much research interest \cite{polson2023, zhao2019, zhao2019a, rapp2024}. Once again, those works focus on the expressive power and augmentation of classical \ac{GP} computations while leaving the idea of quantifying uncertainty to this present work.

%% file: 03-theoretical-background.tex
\section{Theoretical Background}
\label{sec:theo-back}

In the following, we take a probabilistic perspective on the concepts of uncertainty in \acl{ML} before introducing concepts of \acl{QML} to assemble the necessary tools for the proposed mappings from classical to quantum \ac{ML}. 

\subsection{Uncertainty in Machine Learning}
As briefly introduced in \cref{sec:intro}, \ac{UQ} in \ac{ML} characterizes the possibility of estimating the model confidence along with its outputs; this is of special importance for \ac{OOD} data or insufficient knowledge about the data-generating process. 
Generally, two types of uncertainty are distinguished in the area of (Quantum) \ac{ML}, differing in the underlying uncertainty source.
\emph{Aleatoric uncertainty} is intrinsic to the data-generating process and cannot be influenced by more advanced models of the system or a better understanding of the data. \emph{Epistemic uncertainty} assumes insufficient knowledge about the data-generating process and can thus be reduced with additional information about the system. In \cref{sec:exp-setup}, we show how to model these uncertainties in our experiments before evaluating model predictions (visualized as confidence intervals around a learned mean function) in \cref{sec:eval}.

\subsection{Uncertainty Quantification in Classical ML} \label{subsec:theo-UQ-ML}

While many approaches towards \ac{UQ} in classical \ac{ML} exist, we focus on four popular techniques that we later map to the quantum domain. These four techniques (Bayesian modeling, \ac{MC} dropout, Ensemble methods, \acl{GP}es) are introduced in the following.

\subsubsection{Bayesian modeling} \label{subsubsec:theo-UQ-ML-bayesian}

Bayesian modeling serves as the theoretical foundation for \ac{UQ}. The main idea of Bayesian modeling is as follows: Given some training data $\boldsymbol{X}=\left\{\boldsymbol{x}_1,\dots,\boldsymbol{x}_N \right\}$ with associated output values $\boldsymbol{Y}=\left\{\boldsymbol{y}_1,\dots,\boldsymbol{y}_N \right\}$, we want to find model parameters $\boldsymbol{\theta}$ of a model function $\mathbf{f}_{\boldsymbol{\theta}}(\boldsymbol{x})$ that has likely generated the output \cite{galUncertaintyDeepLearning2016}. Using the posterior distribution $p(\boldsymbol{\theta}|\boldsymbol{X},\boldsymbol{Y})$ obtained from Bayes' theorem, we can construct the predictive distribution for some unseen data sample $\boldsymbol{x}^*$ via integrating over all possible parameter configurations as:
\begin{equation}
p(\boldsymbol{y}^*|\boldsymbol{x}^*, \boldsymbol{X}, \boldsymbol{Y})=\int p(\boldsymbol{y}^*|\boldsymbol{x}^* \boldsymbol{\theta})p(\boldsymbol{\theta}|\boldsymbol{X},\boldsymbol{Y})d\boldsymbol{\theta},
\end{equation}
where $p(\boldsymbol{\theta})$ is the prior distribution over the model parameters $\boldsymbol{\theta}$ and the likelihood $p(\boldsymbol{y}^*|\boldsymbol{x}^* \boldsymbol{\theta})$ is commonly chosen to be the softmax likelihood for classification setups and the Gaussian likelihood for regression.
Due to the general intractability of the integral, a popular approximation to this posterior is done via \acl{VI} \cite{jordan1999introduction}, which seeks to minimize the \ac{KL} divergence between a family of parameterized distributions $q_\lambda(\boldsymbol{\theta})$ and the posterior $p(\boldsymbol{\theta}|\boldsymbol{X},\boldsymbol{Y})$ as
\begin{equation} \label{eq:kl}
\lambda^* = \underset{\lambda}{\operatorname{argmin}}\ D_{KL}\left(q_\lambda(\boldsymbol{\theta})\| p(\boldsymbol{\theta}|\boldsymbol{X},\boldsymbol{Y}) \right).
\end{equation}
Optimizing the model parameters using unbiased estimates of gradients of the cost associated with the \ac{KL} divergence leads to the BayesByBackprop algorithm \cite{blundell2015}. Together with the re-parameterization trick, we obtain an easy-to-implement way of obtaining posterior distribution estimates. To further simplify the training optimization, we also generally assume the model functions to come from (independent) Gaussian distributions. In \cref{subsec:mapping-BayesQC}, we see how to map this approach to \ac{QML}.

\subsubsection{\acl{MC} dropout} \label{subsubsec:theo-UQ-ML-mc}

To ameliorate the computational burden of \acl{VI}\footnote{Replacing point estimates with Gaussian mean and variance parameters doubles the number of trainable entities.}, \citeauthor{gal2016} (\citeyear{gal2016}) introduced the method of \ac{MC} dropout. Their approach uses different dropout configurations for every inference pass through the neural network to obtain an ensemble of different (sparse) sub-networks. The obtained \acs{MC} samples are then aggregated to obtain the mean and variance estimators of the probabilistic model.

\subsubsection{Ensemble methods} \label{subsubsec:theo-UQ-ML-ensemble}

Generalizing the notion of different sub-networks above, we can define a \acl{DL} ensemble as an aggregation of several individually trained networks leading to better generalization performance and uncertainty quantification estimates. The individual models can differ in the bootstrap samples of the data set seen during training (bagging) or simply in the initial parameter configuration of the networks \cite{lakshminarayanan2017}. Each individual network can then predict mean and variance estimators related to the input-dependent noise (aleatoric uncertainty) while the aggregation of networks includes the parameter-dependent (epistemic) noise \cite{nemani2023}.

\subsubsection{Gaussian Processes} \label{subsubsec:theo-UQ-ML-gp}

Generalizing the Bayesian viewpoint on neural networks and shifting from inference on latent variables to inference in function space leads to the concept of a \acf{GP}. Formally, a \ac{GP} is a collection of random variables, any finite number of which have a joint Gaussian distribution \cite{carledwardrasmussenGaussianProcessesMachine2004}; as such, we can define a real process $\boldsymbol{f}\colon \mathbb{R}^D \to \mathbb{R}$ with mean function $m(\boldsymbol{x})$ and covariance function $k(\boldsymbol{x}, \boldsymbol{x}')$ as 
\begin{align} \label{eq:gpmeancovar}
    m(\boldsymbol{x}) &= \mathbb{E}\left[ \boldsymbol{f}(\boldsymbol{x}) \right] \quad \text{and}\\
    k(\boldsymbol{x}, \boldsymbol{x}') &= \mathbb{E}\left[ (\boldsymbol{f}(\boldsymbol{x})- m(\boldsymbol{x})) (\boldsymbol{f}(\boldsymbol{x}')- m(\boldsymbol{x}')) \right] \label{eq:gpcovar}
\end{align}
and the corresponding \ac{GP} as  
\begin{equation} \label{eq:gp}
    \boldsymbol{f}(\boldsymbol{x}) \sim \mathcal{GP}\left(m(\boldsymbol{x}), k(\boldsymbol{x}, \boldsymbol{x}') \right)
\end{equation}
where (in the area of \ac{GP} for \ac{ML}) the mean function is often defined to be zero, and the shape of the functions is entirely specified by the covariance, or \textit{\ac{GP} kernel}. 
From this construction, we can then define the posterior (which, as a conditional of multivariate Gaussian distributions, is also a multivariate Gaussian \cite{murphy2022}) as
\begin{align}
f_* | \boldsymbol{X}, \boldsymbol{y}, \boldsymbol{x}_* &\sim \mathcal{N}\left( \overline{f}_*,\ \mathbb{V}\left[ 
 f_* \right] \right) \text{, where} \\
\overline{f}_* &= \mathbf{k}_*^\top\left[ K + \sigma_{\varepsilon}^2 \boldsymbol{I} \right]^{-1} \boldsymbol{y}\quad \text{and} \label{eq:gp:2}\\
\mathbb{V}\left[ f_* \right] &= k(\boldsymbol{x}_*, \boldsymbol{x}_*) - \mathbf{k}_*^\top \left( K + \sigma_{\varepsilon}^2 \boldsymbol{I} \right)^{-1} \mathbf{k}_* , \label{eq:gp:3}
\end{align}
where the kernel matrices $K$, $\mathbf{k}_*$, and $k(\boldsymbol{x}_*, \boldsymbol{x}_*)$ are defined as the covariances of all training points pairs, all training points and a single evaluation point, and the evaluation point with itself, respectively. For a more detailed introduction to necessary \ac{GP} concepts, we refer to \cite{carledwardrasmussenGaussianProcessesMachine2004}.

\subsection{Quantum Machine Learning} \label{subsec:theo-QML}
As the main goal of this paper is to establish mappings from classical to quantum \acs{ML}, we need to introduce some preliminaries of the latter field. In the following, we assume basic knowledge of quantum computing and quantum gates to understand the needed concepts from \acs{QML}; readers not familiar with the necessary prerequisites are referred to \cite{nielsen2010}.

The basic building blocks of \acl{QML} relevant in this paper are variational quantum circuits, which in turn are assembled using parametrized quantum gates. These gates (unitary operators) act on $K$-qubit quantum states represented as $2^K$-dimensional vectors of complex amplitudes. For all experiments in this paper, we focus on single-qubit rotational gates of the form
\begin{equation*}\label{eq:qc-gate}
\begingroup
\renewcommand*{\arraystretch}{1.5}
R(\phi, \gamma, \omega)=\begin{bmatrix*}[r]
e^{-i(\phi+\omega) / 2} \cos (\frac{\gamma}{2}) & -e^{i(\phi-\omega) / 2} \sin (\frac{\gamma}{2}) \\
e^{-i(\phi-\omega) / 2} \sin (\frac{\gamma}{2}) &\  e^{i(\phi+\omega) / 2} \cos (\frac{\gamma}{2})
\end{bmatrix*}
\endgroup
\end{equation*}
with rotational angles $\{\phi, \gamma, \omega \}$. These angles are used to encode the (classical) input into the quantum system by unitary operators
\begin{equation} \label{eq:rot-gate}
    U_j(\boldsymbol{x};\ \boldsymbol{w}_j, \boldsymbol{b}_j) = R\left(\boldsymbol{w}_j \circ \boldsymbol{x} + \boldsymbol{b}_j \right)
\end{equation}
for $\boldsymbol{x}\in \mathbb{R}^3$, where the Hadamard-product parameter $\boldsymbol{w}_j\in \mathbb{R}^3$ and the additive parameter $\boldsymbol{b}_j \in \mathbb{R}^3$ are optimized during training.
We can stack these variational gates arbitrarily for general $\boldsymbol{x}\in \mathbb{R}^N$ (hence the index $j$). This way, each input feature can be encoded an arbitrary number of times, accordingly called data re-upload encoding \cite{perez-salinas2019}. The second type of gate structure needed in the following is the two-dimensional Controlled-NOT (CNOT) gate used to entangle quantum states. Together with a final measurement process, the resulting \acs{QML} model function can be written as
\begin{align}
    f(\boldsymbol{x} ; \boldsymbol{\theta}) &=\left\langle 0\left|U(\boldsymbol{x};\boldsymbol{\theta})^{\dagger} \mathcal{M} U(\boldsymbol{x}; \boldsymbol{\theta})\right| 0\right\rangle  \label{eq:qml-func:1} \\
    &=\sum_{\boldsymbol{\omega} \in \Omega} c_{\boldsymbol{\omega}} e^{i \boldsymbol{\omega} \boldsymbol{x}} \label{eq:qml-func:2}
\end{align}
where \cref{eq:qml-func:1} captures the parameterized quantum circuit $U(\boldsymbol{x}; \boldsymbol{\theta})$ acting on the initial zero state $|0\rangle$ and the evaluated expectation value of the measurement observable $\mathcal{M}$. The importance of \cref{eq:qml-func:2} lies in the ability to write parameterized quantum circuit functions as a Fourier series whose frequency spectrum $\Omega \subset \mathbb{R}^N$ is entirely defined by the input encoding and whose coefficients $c_{\boldsymbol{\omega}}$ depend on the design of the circuit architecture \cite{schuld2021}. This notation helps us to understand the idea of quantum kernels when discussing quantum \acl{GP}es in \cref{subsec:mapping-gps}.

%% file: 04-mapping-ml-qml.tex
\section{Mapping \acs{UQ} from \acs{ML} to \acs{QML}}
\label{sec:mapping-uq-qml}
Given the preliminaries on \acl{QML} and \acl{UQ} outlined above, we now turn to the combination of these fields for mapping classical \ac{UQ} to variational quantum circuits.

\subsection{Bayesian Quantum Machine Learning} \label{subsec:mapping-BayesQC}

As discussed in \cref{sec:related-work}, first steps in the direction of Bayesian quantum computing are taken in \cite{nguyen2022}, which are described in the following. 

The essential idea is the same as in the classical BayesByBackprop algorithm: As introduced in \cref{subsubsec:theo-UQ-ML-bayesian}, we are interested in the minimization of the KL divergence from \cref{eq:kl}, with the additional assumption of diagonal Gaussian distributions for the variational posterior \cite{nguyen2022}. For deterministic circuit parameters $\boldsymbol{\theta}=\{\boldsymbol{\mu}, \boldsymbol{\rho} \}$ and parameter-free noise $\boldsymbol{\epsilon}$, we arrive at a parameterized Gaussian distribution as $\mathcal{N}\left( \boldsymbol{\mu}, \boldsymbol{\sigma}^2= \left[ \operatorname{log}(1+ \operatorname{exp}(\boldsymbol{\rho})) \right]^2 \right)$ and resulting posterior weight samples as
\begin{equation} \label{eq:post-weights}
    \boldsymbol{w} = \boldsymbol{\mu} + \operatorname{log}(1+ \operatorname{exp}(\boldsymbol{\rho})) \circ \boldsymbol{\epsilon}
\end{equation}
where instead of classical network weights, we use the rotational angles of the quantum circuits as a basis for the construction of the variational model functions. This approach will become clear when introducing the specific quantum circuit architectures in \cref{subsec:qml-architectures}.

Thus, to obtain a Bayesian \acl{QML} model, we iteratively evaluate the loss function 
\begin{align}
    &\mathcal{L}\left(\boldsymbol{x}_i, y_i, \boldsymbol{\theta} \right) =  \mathcal{L}_{KL} +  \mathcal{L}_{obj},\ \text{for} \label{eq:bayes-loss-1}\\
    &\mathcal{L}_{KL}  = \frac{1}{2} \sum_{k=1}^{K} \left(1 + \log ((\sigma_{k})^2) - (\mu_{k})^2 - (\sigma_{k})^2  \right), \label{eq:bayes-loss-2} \\
    &\mathcal{L}_{obj} = \mathcal{L}_{MSE} \quad \text{or} \quad \mathcal{L}_{CE}
\end{align}
for multiple posterior weight samples from \cref{eq:post-weights} to optimize the \acs{QML} model parameters $\boldsymbol{\theta}$. The index $k$ in \cref{eq:bayes-loss-2} defines the $k$-th element of the mean and variance vectors coming from the Gaussian prior and variational posterior \cite{kingma2013}. As discussed in \cref{subsubsec:theo-UQ-ML-bayesian}, we typically choose the Mean-Squared-Error (MSE) function for Gaussian likelihood objectives in regression or the Cross-Entropy (CE) for softmax likelihood classification tasks.

\subsection{Quantum \acl{MC} Dropout} \label{subsec:mapping-MC-dropout}

In classical \ac{ML}, the common choice to perform dropout is to randomly select some nodes in all intermediate layers and set their output to zero. Mathematically, we can write this approach for a typical hidden layer function $y_{k}=\sigma \left( \boldsymbol{W}_k \ \boldsymbol{y_{k-1}} + \boldsymbol{b}_k\right)$ of layer $k$ (and non-linearity $\sigma$) as a Hadamard product of the output with some Bernoulli vector as $\hat{y}_{k} = y_{k} \circ z_k$, where for each element $z_k^{(i)}$ of $z_k$ it holds $z_k^{(i)}\sim \operatorname{Bernoulli}(p)$, given a dropout probability of $p$. While this approach is the most popular, many different dropout methods exist, e.g., using  $z_k^{(i)}\sim \mathcal{N}(1,\sqrt{\frac{p}{1-p}})$ for Gaussian dropout.

In a quantum circuit architecture consisting of rotational and entangling gates, we can think of similar options to introduce dropout; however, the resulting impact on the model function differs.
While classical neural network structures commonly have multiple information streams from the input to the output nodes, the quantum state on which the circuit acts is influenced and modified with each dropout operation. Thus, the modification of a single gate in the circuit (e.g., dropping a rotational gate or modifying its variational parameters) changes all amplitudes of the quantum state vector, which in turn influences the measurement and observed model output. 

In this light, we propose three different techniques -- varying in the impact of the quantum state amplitudes -- that are comparatively evaluated in \cref{sec:eval}, along with the other probabilistic \acs{QML} models described in this work. These three techniques are as follows: 
First, we only drop additive parameters $\boldsymbol{b}_j$ from the rotational gates as defined in \cref{eq:rot-gate}. Second, we drop whole rotation gates from the quantum circuit by replacing the specified gates with identity operators in the circuit. Last, we investigate the influence of Gaussian noise to the model parameters during each forward pass in the quantum circuit, equivalently to classical Gaussian dropout. The last variant is of particular interest as it can be represented by imperfect rotations of quantum states on the Bloch sphere, arising naturally from noise in the quantum system. For the final \acs{UQ} model output, multiple forward passes through the quantum circuit are aggregated as done in classical \ac{MC} dropout, to obtain
\begin{align}
    \boldsymbol{\mu} &= \frac{1}{M}\sum_{m=1}^{M} \hat{\boldsymbol{y}}_m \quad \text{and} \label{eq-dropout-agg1}\\
    \boldsymbol{\sigma} &= \sqrt{\frac{1}{M-1}\sum_{m=1}^{M} (\hat{\boldsymbol{y}}_m - \boldsymbol{\mu})^2}, \label{eq-dropout-agg2}
\end{align}
where $M$ inference evaluations (each producing an individual model output $\hat{\boldsymbol{y}}_m$) are averaged for empirical mean and standard deviation estimators.

\subsection{Quantum Circuit Ensembles} \label{subsec:mapping-ensembles}

The mechanism of using an aggregation of different models in an ensemble as described in Subsection \ref{subsubsec:theo-UQ-ML-ensemble} can be easily extended to \ac{QML}, in which case multiple circuit outputs are aggregated to obtain an approximation of the posterior predictive distribution. First investigations on quantum ensembles explore the possibility of parallelism to evaluate predictions of exponentially large ensembles \cite{schuld2017} and the quantum Bayesian approximation using deep quantum ensembles \cite{nguyen2022}. However, evaluating uncertainty estimates of quantum ensembles is a gap filled in the following.

As noted in \cref{subsec:theo-QML}, the family of functions that a \acs{QML} model with re-upload encoding can learn is characterized by a partial Fourier series \cite{schuld2021}, as formalized in \cref{eq:qml-func:2}. Thus, all quantum circuit functions investigated in the following are periodic in nature, a property that shapes the predicted mean functions as well as the estimated confidence intervals.
As done in \cref{subsec:mapping-MC-dropout}, we use the empirical mean and standard deviation to aggregate the model results and obtain an approximation for the predictive posterior (see Equations \eqref{eq-dropout-agg1} and \eqref{eq-dropout-agg2}) for an ensemble consisting of $M$ models, each generating an individual model output $\hat{\boldsymbol{y}}_m$.

\subsection{Gaussian Processes in \acl{QML}} \label{subsec:mapping-gps}

As seen in \cref{subsubsec:theo-UQ-ML-gp}, the notion of a kernel -- defined by the covariance of the random variables within the \ac{GP} and hence influencing the overall shape of the predictive functions -- lies at the core of a \acl{GP}. Thus, we now describe how research combines \acs{GP}s with \emph{quantum kernels}.

Quantum kernels are omnipresent in the domain of \ac{QML}, as they arise naturally from the feature map of the encoding of classical information into the quantum state of the system, i.e., the embedding in the (quantum) Hilbert space \cite{schuld2019a}.

In general, any feature map $\phi : \mathcal{X} \to \mathcal{F}$ from an input space $\mathcal{X}$ to some feature space $\mathcal{F}$ paired with the definition of an inner product on $\mathcal{F}$ gives rise to a kernel as
\begin{equation}
    k(x, x') := \langle \phi (x), \phi (x') \rangle_{\mathcal{F}}.
\end{equation}
With the help of quantum circuits, we can make use of this definition and replace the classical feature map $\phi (x)$ with a quantum embedding $\ket{\phi(x)}$. The resulting quantum kernel can be written as 
\begin{equation}
    k(x,x')= \left|\braket{\phi(x')|\phi(x)}\right|^2 = \left|\bra{0} U(x')^\dag U(x)\ket{0} \right|^2
\end{equation}
as shown in \cite{rapp2024}. For the chosen re-upload encoding method in this paper, we can write the quantum kernel as
\begin{equation} \label{eq:reup-kernel}
    k(\mathbf{x},\mathbf{x}')= \sum_{\mathbf{s}, \mathbf{t}\in \Omega} e^{i \mathbf{s}\mathbf{x}}e^{i\mathbf{t}\mathbf{x}'}c_{\mathbf{s}\mathbf{t}}
\end{equation}
for $\Omega \subseteq \mathbb{R}^n$ and $c_{\mathbf{s}\mathbf{t}} \in \mathbb{C}$ \cite{schuld2021a}, where for ease of notation we consider a single non-trainable upload of the data. Comparing this result to \cref{eq:qml-func:2}, it is easy to see that the obtained kernel from \cref{eq:reup-kernel} is a direct consequence of the Fourier representation of the chosen quantum circuit architecture and offers an elegant way to provide a rich feature space despite the low complexity associated with the circuit structure.

%% file: 05-experimental-setup.tex
\section{Experimental Setup}
\label{sec:exp-setup}

In this section, we briefly mention the setup used for the empirical experiments in the following \cref{sec:eval}, including the datasets, the model architectures, and the general optimization and evaluation workflow. This allows us to have a structural baseline for both the experimental data and the architectural design of the circuits.

\subsection{Datasets and Pre-Processing} \label{subsec:datasets}

\begin{figure}[htbp]
\vskip 0.1in
\begin{center}
\centerline{\includegraphics[width=\columnwidth]{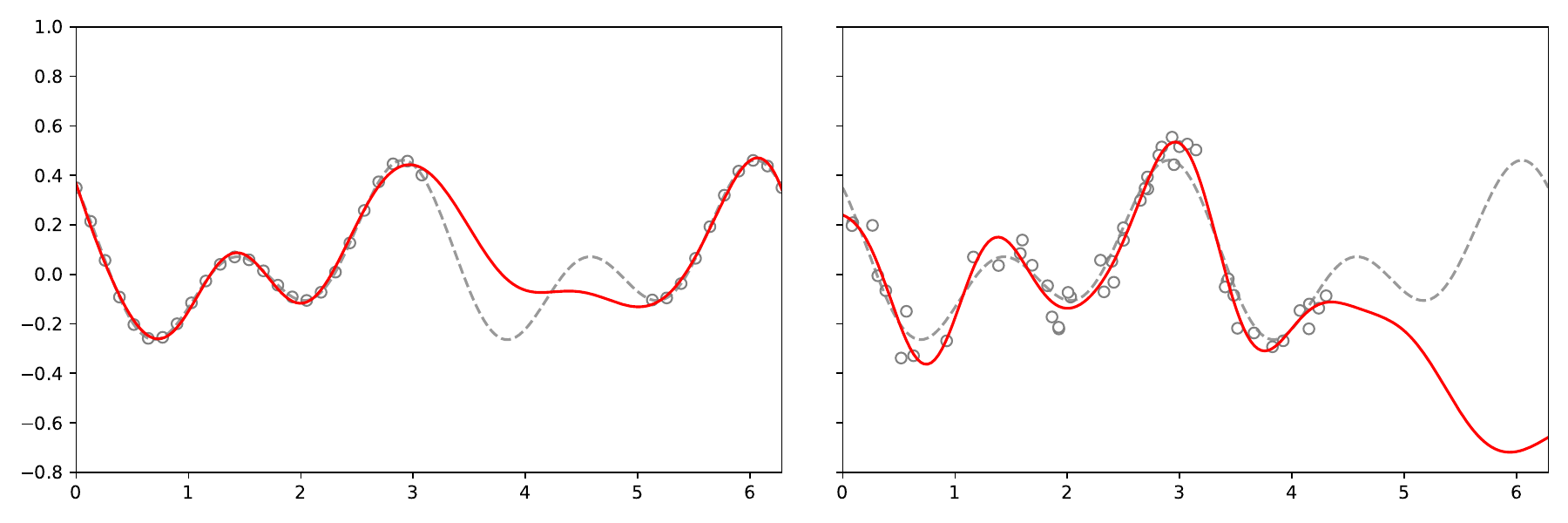}}
\caption{Regression functions representing epistemic-only uncertainty (left) and a combination of aleatoric and epistemic uncertainty (right). In red, the output of a deterministic \acs{QML} model is given, showing its inability to adhere to the correct periodicity of the underlying data function when noise is present in the system.}
\label{fig:regression-data}
\end{center}
\vskip -0.2in
\end{figure}

All models are trained and evaluated on a regression task in the main text and a classification task in Appendix \ref{app:class}. Both datasets are kept low-dimensional to permit stable training and descriptive visualizations of uncertainty estimates. For regression, we adopt the dataset from \citeauthor{schuld2021} (\citeyear{schuld2021}), which consists of a one-dimensional Fourier series of degree two -- naturally fitting the set of functions a quantum circuit can learn perfectly -- as given by
\begin{equation}
    f(x) = \sum_{n=-2}^{2} c_n e^{-inx}
\end{equation}
where the coefficients $c_n$ allow us to modify the shape of the function. For the experiments, we use the configuration $c_0 = 0.1,\ c_1= c_2 = 0.15- 0.15i$ and the fact that $c_{-k} = c^*_{k}$, leading to the function shown as a dashed line in \cref{fig:regression-data}.

To estimate the uncertainty of the models, we construct two adapted versions of this regression setup: First, we include a gap for $x_i\in [\pi,\ 1.6\pi)$ (epistemic uncertainty). Second, we include a gap for $x_i\in [1.4\pi,\ 2\pi)$ and sample from the modified function
\begin{equation}
    f_\varepsilon(x) = f(x) + \varepsilon \quad \text{for}\ \varepsilon \sim \mathcal{N}(0,\sigma_\varepsilon^2)
\end{equation}
with $\sigma_\varepsilon=0.1$ to include aleatoric uncertainty. Both versions of the dataset are visualized in \cref{fig:regression-data}, where the individual training data points are depicted as gray circles.

For classification, the popular two-moons\footnote{\url{https://scikit-learn.org/stable/modules/generated/sklearn.datasets.make_moons.html}} dataset serves as the benchmark for empirical experiments as it enables clear visualization of the non-linear decision boundary while being low-dimensional enough to work within the \ac{QML} regime. See Appendix \ref{app:class} for all classification experiments.

\subsection{Quantum Machine Learning Architectures}\label{subsec:qml-architectures}
\begin{figure}[htbp]
\vskip 0.1in
\begin{center}
\centerline{\includegraphics[width=\columnwidth]{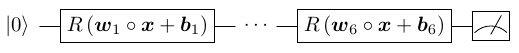}}
\caption{The basic architecture for all quantum models, on which the different mappings from classical to quantum \acs{UQ} are developed and evaluated.}
\label{fig:qml-circuit}
\end{center}
\vskip -0.2in
\end{figure}

Although the individual mappings from classical to quantum \acs{UQ} differ dramatically by design, where, e.g., ensemble methods are a mixture of different base models and \ac{MC} dropout employs a single model with different dropout configurations (and \ac{GP} models have a different basis for modeling the data distribution altogether), we try to keep a similar structure for all (base-) models. Doing so allows us to compare the uncertainty estimates more effectively, using a common structural ground. This basic architecture consists of six rotational layers in a single-qubit quantum circuit followed by a calculation of the expected value of a measurement in the Pauli-Z basis\footnote{This means that the \ac{QML} model output is bounded by $f(\boldsymbol{x};\boldsymbol{\theta}) \in [-1,1]$ by definition of expectation values \cite{nielsen2010}.}. The circuit is given in \cref{fig:qml-circuit}.
To adjust the univariate data input to the rotational parameter vectors, we stack the input into three-tuples as $\boldsymbol{x}=\left(x, x, x \right)$. 
The main idea behind this architecture is the relationship to the proposed regression and classification tasks, where the chosen model design showed a good trade-off between the ability to fit target functions despite the use of dropout (or other \acs{UQ} techniques) and possible overfitting scenarios.

The training procedure of the models follows best practices from classical and quantum \acs{ML}, where the described quantum circuit ansatz is optimized using the Adam optimization scheme \cite{kingma2017} with a learning rate of $\eta=0.01$, an upper bound of $1000$ training epochs and early stopping cutoffs at $\mathcal{L}_{MSE} < 0.005$ for regression and $\mathcal{L}_{CE} < 0.3$ for classification.
More details about the model architecture, optimization process, and evaluation can be found in Appendix \ref{app:opt-eval}.

%% file: 06-empirical-evaluations.tex
\section{Empirical Evaluation}
\label{sec:eval}
With the tools assembled throughout this work, we now describe the outcomes of applying \ac{UQ} techniques from \cref{sec:mapping-uq-qml} to typical \ac{QML} model architectures as constructed in \cref{sec:exp-setup}. After qualitative empirical results for each model are shown, we introduce the measure of calibration error and comparatively evaluate all architectures.

\subsection{Qualitative Analysis} \label{subsec:eval-qualitative}
For qualitative analysis, we visualize the predictive output distribution for each of the proposed quantum \acs{UQ} models, including a Bayesian quantum circuit, three \ac{MC} dropout models (bias-only, rotation, Gaussian dropout), a quantum circuit ensemble and a \ac{GP} equipped with a quantum kernel. In each case, we plot the mean output function and shade the area containing one and two standard deviations, respectively.
\begin{figure}[htbp]
\vskip 0.1in
\begin{center}
\centerline{\includegraphics[width=\columnwidth]{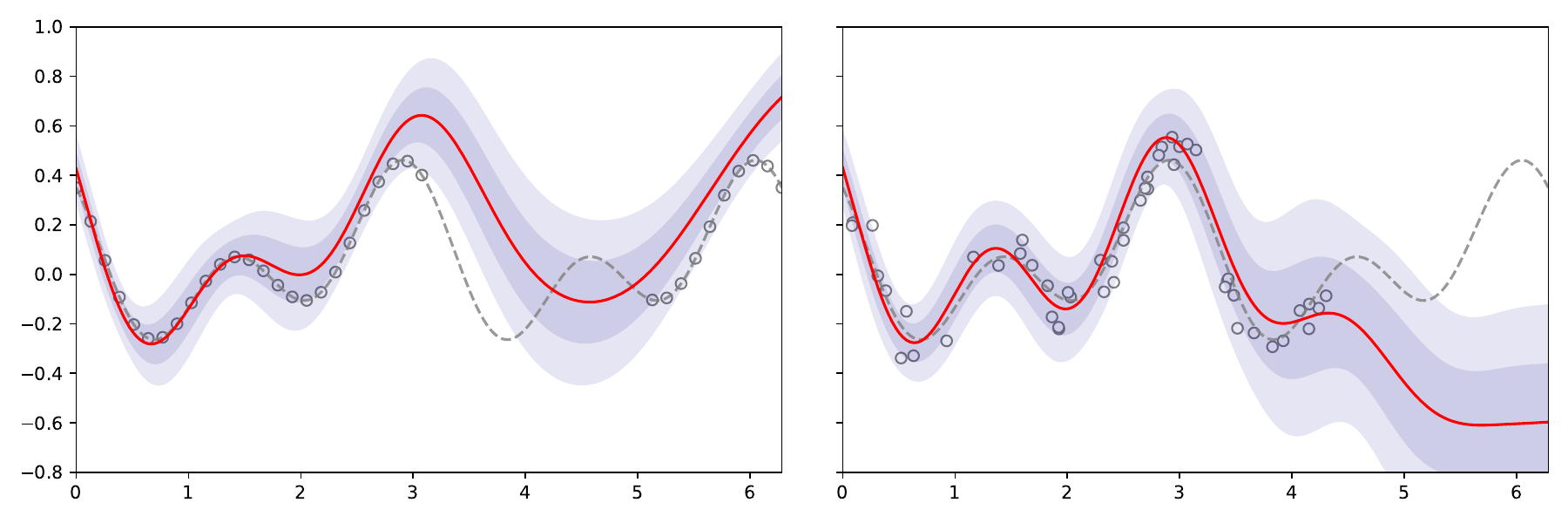}}
\caption{Results for a Bayesian \ac{QML} model for a noiseless (left) and a noisy regression task (right).}
\label{fig:regress-bayesianQML}
\end{center}
\vskip -0.2in
\end{figure}

First, we show the results for Bayesian \acs{QML} regression, where we use the six-layer architecture in combination with the Bayesian framework and the re-parameterization trick as described in \cref{subsec:mapping-BayesQC}. \cref{fig:regress-bayesianQML} shows the resulting model outputs for both regression scenarios (epistemic only versus epistemic and aleatoric uncertainty). In both cases, the model cannot fully capture the function shape once data samples are missing but reflects this uncertainty in larger deviation bounds within the respective regions. While epistemic uncertainty is reflected in the predictive variance estimators, the posterior-pushforward modeling \cite{nemani2023} cannot truthfully capture aleatoric uncertainty.
\begin{figure}[htbp]
\vskip 0.1in
\begin{center}
\centerline{\includegraphics[width=\columnwidth]{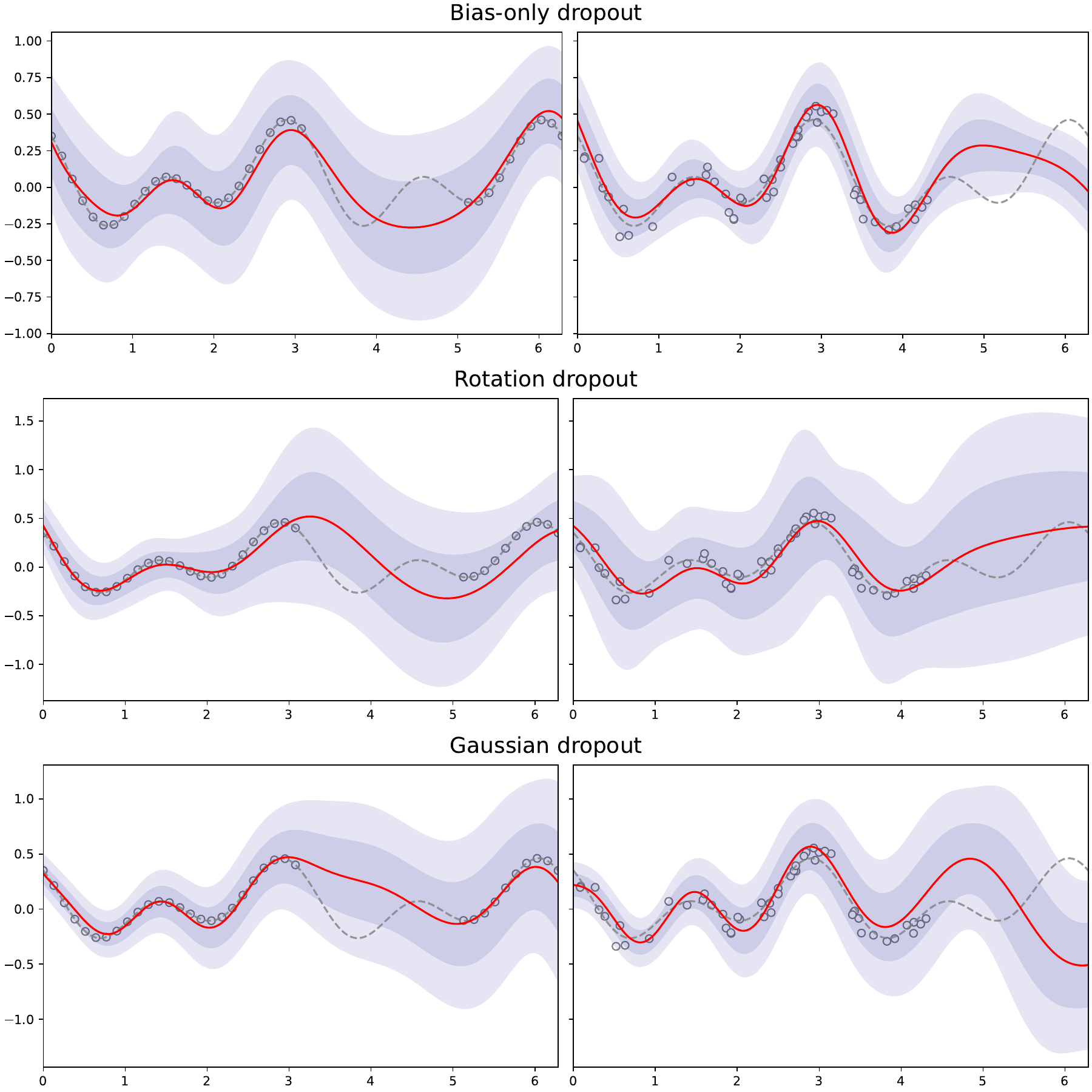}}
\caption{Regression results for all \ac{MC} dropout \ac{QML} models. }
\label{fig:regress-mc}
\end{center}
\vskip -0.2in
\end{figure}

Next, we investigate quantum \acs{MC} dropout by evaluating all chosen dropout configurations (bias-only, rotation, Gaussian dropout) as shown in \cref{fig:regress-mc}. More precisely, we use the same architecture as before and apply a dropout rate of $p=0.1$ to the dropout algorithms described in \cref{subsec:mapping-MC-dropout}.
While all models show uncertainty in regions of training data absence, the models differ in their predicted variance: Bias dropout leads to relatively tight uncertainty bounds, especially in the noisy regression setting. Rotation dropout poses a more significant structural burden for each inference pass, leading to larger confidence intervals. Gaussian dropout shows the best visual results, given the slight changes in rotation angles due to Gaussian noise. As before, the predictive distributions only capture epistemic uncertainty.
\begin{figure}[htbp]
\vskip 0.1in
\begin{center}
\centerline{\includegraphics[width=\columnwidth]{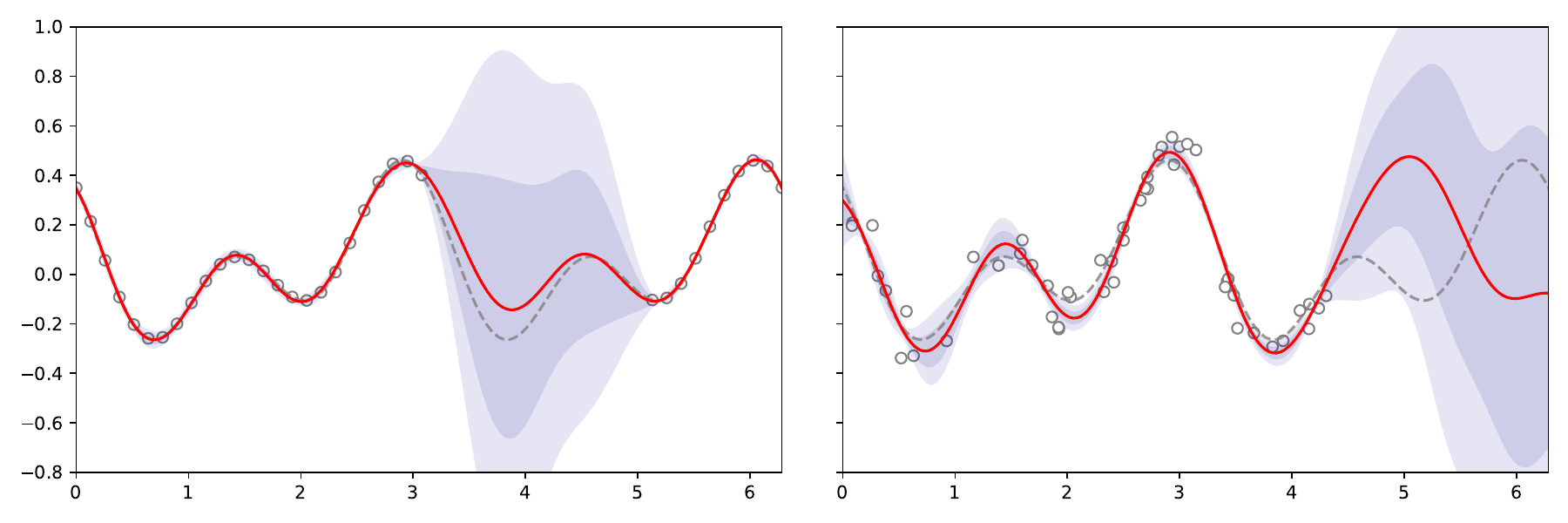}}
\caption{Regression results for an ensemble of \ac{QML} models.}
\label{fig:regress-ensemble}
\end{center}
\vskip -0.2in
\end{figure}

For ensembles, this behavior changes. When aggregating eight individual \acs{QML} model outputs to form mean and standard deviation estimators, the ability to represent epistemic and aleatoric uncertainty emerges, as seen in \cref{fig:regress-ensemble}. Intuitively, the confidence shapes align well with the underlying problem statement, where a near-perfect fit is achieved in regions of sufficient data, and significant deviations from the mean function emerge in the absence of data samples.

\begin{figure}[htbp]
\vskip 0.1in
\begin{center}
\centerline{\includegraphics[width=\columnwidth]{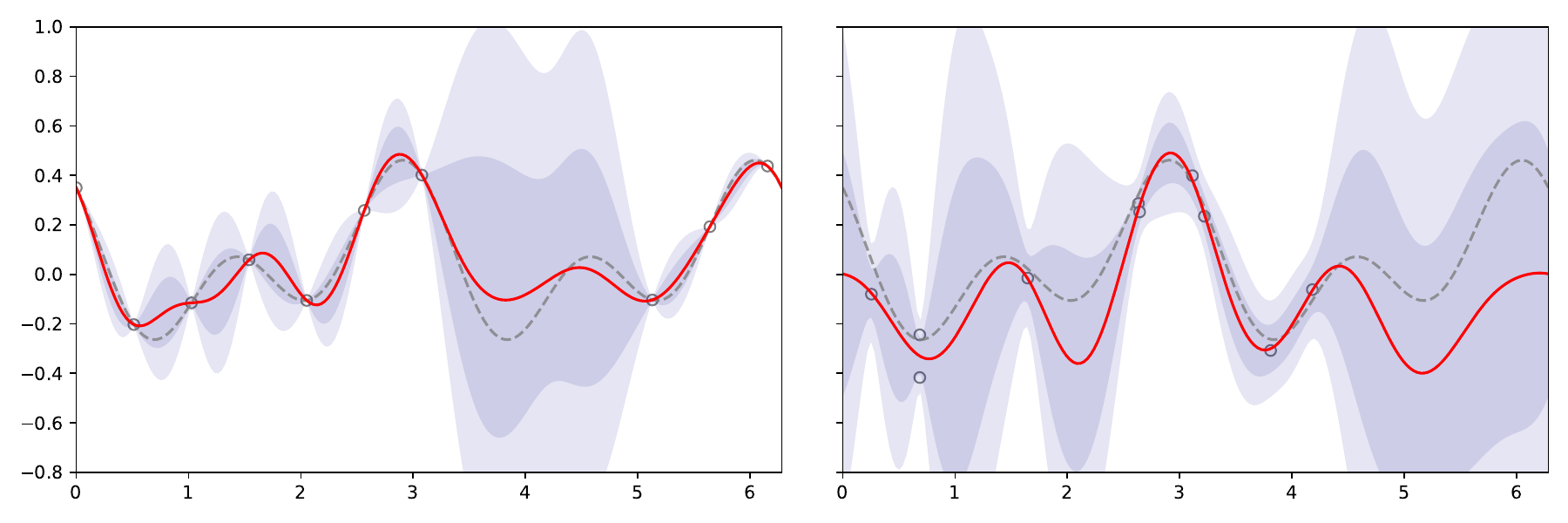}}
\caption{Gaussian Process regression results using a quantum re-upload encoding kernel consisting of two qubits and two layers. }
\label{fig:regress-gp}
\end{center}
\vskip -0.2in
\end{figure}

Last, we study the uncertainty estimates given by \acl{GP} regression using a quantum kernel, as visualized in \cref{fig:regress-gp}. The mathematical structure of the \ac{GP} allows perfectly capturing noiseless data samples (left subfigure) and adapting to aleatoric uncertainty by incorporating the noise into the covariance matrices. 
The high-frequency structure of the employed quantum kernel leads to relatively large uncertainty bounds between samples, even in regions of higher data density. The uncertainty shapes associated with different underlying kernel circuit structures are investigated in Appendix \ref{app:gp-diff-kernels}.

\subsection{Quantitative Analysis} \label{subsec:eval-quantitative}
\begin{figure}[htbp]
\vskip 0.1in
\begin{center}
\centerline{\includegraphics[width=\columnwidth]{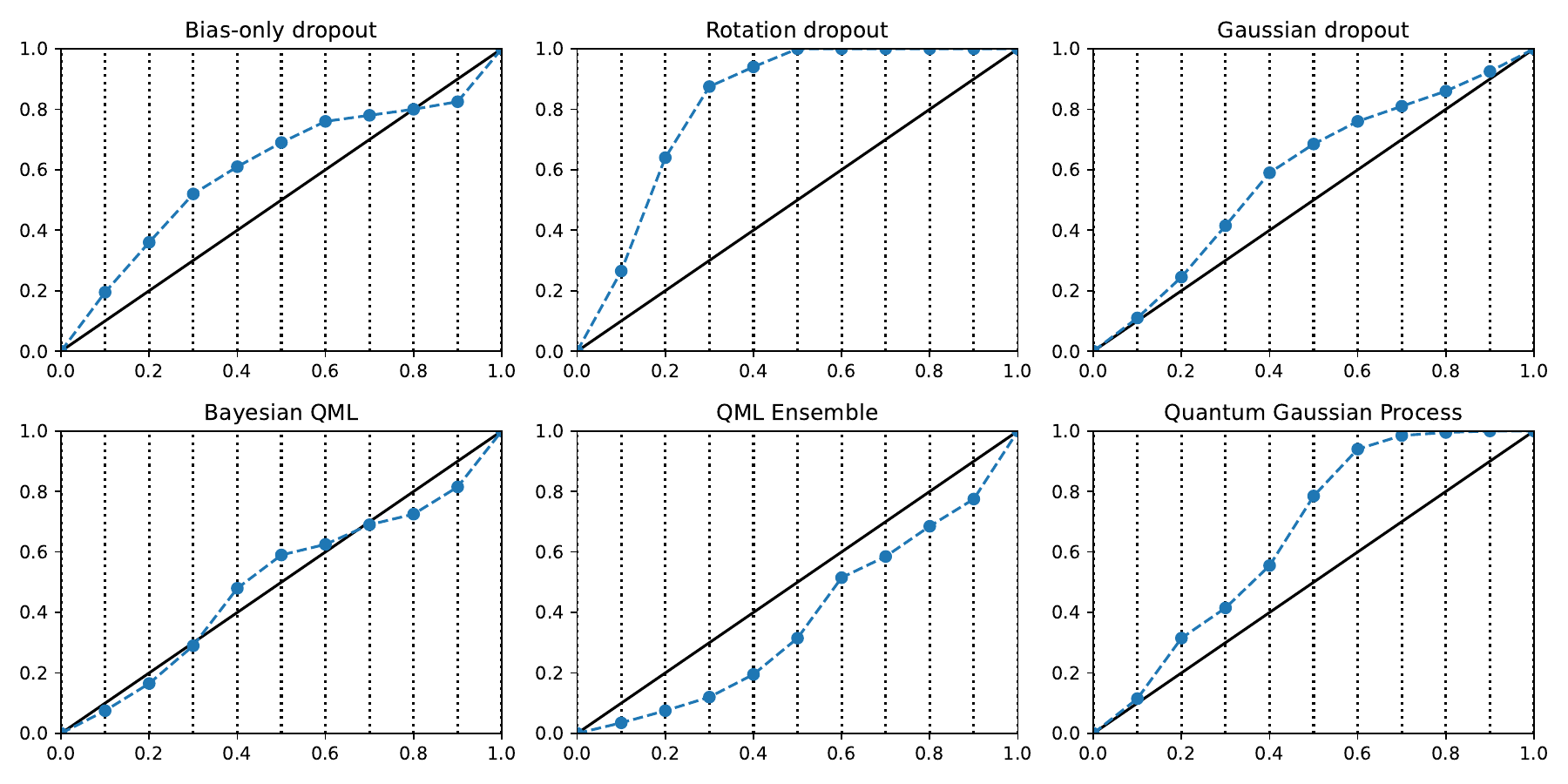}}
\caption{Calibration curves for all models investigated in this paper. Large deviations from the diagonal indicate poor calibration, where lines over (under) the diagonal signal underconfidence (overconfidence).}
\label{fig:ece-all}
\end{center}
\vskip -0.2in
\end{figure}

For a more rigorous evaluation, we introduce the notation of the \emph{\acf{ECE}} \cite{pakdamannaeini2015} as the expected difference between the model's confidence and accuracy. If we plot the observed confidence (actual model output accuracy within some confidence interval) on the y-axis against the expected confidence (expected number of samples to lie within that interval) on the x-axis, we obtain the calibration curves as shown in \cref{fig:ece-all}, where larger deviations from the diagonal express larger miscalibration \cite{kuleshov2018}.
To obtain the \acs{ECE}, we then take the mean absolute distance between the curves for each of the chosen confidence intervals (in \cref{fig:ece-all}, there are 11 confidence intervals, ranging from $0$ to $100\%$). Together with insights from individual model implementations, we summarize the results from the quantitative analysis in \cref{tab:uq-quality}.

First, we note the mere possibility of mapping \ac{UQ} methods from classical \ac{ML} to \ac{QML}. Since the underlying model functions differ between classical and quantum \ac{ML} and we have a much more complex embedding space when working on qubits instead of classical bits, it is by no means trivial to infer that the chosen classical techniques are applicable in \ac{QML} as well as they proved to be. Even more so, given the empirical evaluation, it becomes clear that different methods have different properties of computational requirements, quality of uncertainty predictions, and ease of integration into existing \ac{QML} architectures. Out of all models, the Bayesian \acs{QML} framework shows the best calibration; however, during the empirical evaluations, we found that even slight modifications of circuits, such as Gaussian noise influences for quantum \acs{MC} dropout, results in well-calibrated \acs{QML} models with a trivial implementation overhead. Our implementation of quantum ensembles was the only technique that shows \emph{overconfidence} in the model predictions. In contrast, dropout and \acs{GP} models are slightly \emph{underconfident}. In safety-critical scenarios, this behavior is better aligned with the intention of uncertainty quantification; however, we find that in any case, the underlying circuit architecture needs to be adapted to the specific use-case.
\begin{table}[htbp]
  \renewcommand{\arraystretch}{1.2}
  \renewcommand\tabularxcolumn[1]{m{#1}}
  \newcolumntype{Y}{>{\centering\arraybackslash}X}
  \caption[Uncertainty Evaluation]{Findings of comparative evaluation for all models, including the \acs{ECE}, computational cost, and quality of uncertainty.} \label{tab:uq-quality}
  \centering
  \vskip 0.15in
  \begin{tabularx}{\columnwidth}{| l || c | Y | Y |}
  \hline
    \textbf{Model} & \textbf{\ac{ECE}} & \textbf{Cost} & \textbf{Quality} \\
    \hline 
    \hline
    Bayesian \ac{QML}& $0.0395$ & Medium & High\\
    \hline
    Bias-only \ac{MC} d. & $0.1082$ & \multirow{3}{*}{Low} & \multirow{3}{*}{Low-Med.} \\
    \cline{1-2}
    Rotation \  \ac{MC} d. & $0.2927$ && \\
    \cline{1-2}
    Gaussian \ac{MC} d. & $0.0818$ && \\
    \hline
    \ac{QML} Ens. & $0.1091$ & Medium &  Medium\\
    \hline
    Quantum GPs & $0.1459$ & High & Medium \\
    \hline
  \end{tabularx}
\end{table}

%% file: 10-summary-outlook.tex
\section{Summary and Outlook}
\label{sec:summary}
In this work, we extend foundations on quantum Bayesian modeling, quantum kernels, and ensemble theory to the applications of quantum \acs{UQ}. Furthermore, we develop new ideas of bridging \ac{MC} dropout techniques to \ac{QML}, building upon their potential in classical \acs{ML}. The resulting model calibration is comparatively evaluated for all proposed approaches, first qualitatively via deviation bounds and quantitatively via calibration metrics.

The empirical results indicate the potential of \acs{UQ} in \acs{QML}. Most strikingly, a Bayesian approach to \acs{QML} model optimization (or a similar noise-centric viewpoint via Gaussian dropout) provides a good trade-off between implementation complexity and quality of uncertainty prediction. While we shed light on multiple possible extensions at the intersection of \acs{UQ} and \acs{QML}, many avenues are yet to be explored, where we mention the extension to more complex scenarios as well as the inclusion of further \acs{UQ} techniques here and give a more detailed overview of limitations and future work in Appendix \ref{app:limit}.

Incorporating insights into the field of \acl{UQ} in the context of \acs{QML} is an important -- but often overlooked -- step when developing quantum algorithms, as we see quantum hardware become increasingly powerful and quantum algorithms increasingly complex. In line with the principle of ``security by design,'' we should therefore integrate trustworthiness into \acs{QML} from the outset to avoid encountering the same issues as faced in classical \acs{ML} approaches.

%% file: 99-appendix.tex
\appendices
\section{Model Optimization and Evaluation} \label{app:opt-eval}

As stated in the main text, the basic optimization and evaluation procedure used in the empirical evaluations in \cref{sec:eval} are designed following best practices from classical and quantum \ac{ML}. First, the described quantum circuit ansatz consisting of six re-upload encoding layers and one qubit is evaluated on a set of training points to compute the value of a chosen loss function. For regression experiments, we make use of the mean-squared error function\footnote{\url{https://pytorch.org/docs/stable/generated/torch.nn.MSELoss}}, whereas the binary cross-entropy function\footnote{\url{https://pytorch.org/docs/stable/generated/torch.nn.BCEWithLogitsLoss}} serves as a basis for classification optimization. The gradient of the chosen loss function is then used to update the model parameters, using the Adam \cite{kingma2017} algorithm, where a learning rate of $\eta =0.01$ is used for standard model optimization. For models that use aggregation, such as dropout or ensemble circuits, we use ten forward passes through the model to estimate the predictive mean in each training step and 1000 model evaluations on the test set to produce the figures in \cref{subsec:eval-qualitative}. For the Bayesian treatment of \ac{QML} using BayesByBackProp, we substitute the standard loss functions with \cref{eq:bayes-loss-1} as derived in \cref{subsec:mapping-BayesQC}. For the \acs{GP}, it was found during training, that the basic one-qubit six-layer architecture leads to instabilities during training, and we switched the underlying quantum circuit kernel to an architecture consisting of two qubits and two layers (see Appendix\ref{app:gp-diff-kernels}).

All models are trained until convergence on the regression and classification datasets. That is, we have an upper bound of 1000 epochs for all models, with an early stopping cutoff if the mean-squared error loss falls below a threshold of $\mathcal{L}_{MSE} < 0.005$ for regression or if the binary cross-entropy is $\mathcal{L}_{CE} < 0.3$ for classification.

The quantum model and optimization routines are implemented on a PyTorch \cite{paszke2019} state-vector simulator based on the Pennylane \cite{bergholm2022} \textit{StronglyEntanglingLayers}\footnote{\url{https://docs.pennylane.ai/en/stable/code/api/pennylane.StronglyEntanglingLayers}} template for the possibility of integration into existing \ac{ML} pipelines and efficient gradient evaluation. Note that no entanglement is required for one qubit, and the StronglyEntanglingLayers template collapses to repeated rotational layers where we incorporate the input for re-upload encoding, as shown in \cref{eq:rot-gate}. For the quantum kernel \ac{GP} algorithms, we utilize the scikit-learn library\footnote{\url{https://scikit-learn.org/stable/modules/gaussian_process}}.

\section{Gaussian Process Regression for Different Re-Uploading Configurations} \label{app:gp-diff-kernels}

During the experiments with quantum \ac{GP} regression in \cref{subsec:eval-qualitative}, we noted the dependency of the resulting output functions on the underlying kernel. More specifically, we obtained drastically different mean function shapes and confidence intervals in the evaluations by changing the number of re-upload encoding layers for our chosen circuit architecture.

\begin{figure}[ht!]
\centering
\begin{subfigure}[b]{0.95\columnwidth}
   \includegraphics[width=\columnwidth]{regress_gaussian_process.pdf}
   \caption{Quantum \acl{GP} regression output for shallow circuit kernel.}
   \label{fig:gpshort} 
\end{subfigure}

\begin{subfigure}[b]{0.95\columnwidth}
   \includegraphics[width=\columnwidth]{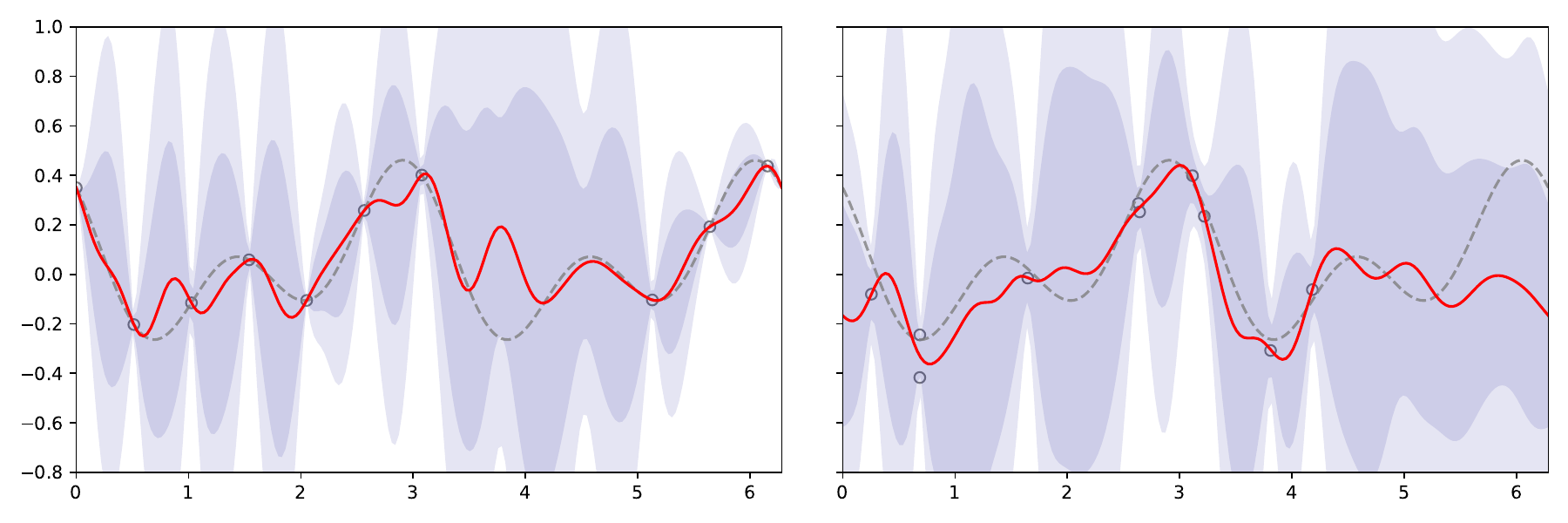}
   \caption{Quantum \acl{GP} regression output for deep circuit kernel.}
   \label{fig:gplong}
\end{subfigure}
\caption[Quantum \acl{GP} regression for different circuit kernel lengths.]{Comparative overview over regression output functions showing the influence of the underlying circuit complexity.}
\label{fig:gp-different-circuit-lengths}
\end{figure}

As a baseline, we again visualize the quantum \ac{GP} regression plots (here: \cref{fig:gpshort}) from the empirical evaluation, where we employ a two-qubit, two-layer quantum circuit architecture. 

If we increase the number of layers to obtain a two-qubit, six-layer case (which is still relatively low-scale compared to typical circuit architectures in the range of $10-10^3$ layers), we obtain the \ac{GP} model output depicted in \cref{fig:gplong}. We can see the difference in the mean function, showing a behavior typical for overfitting, with large fluctuations in the intervals between data points. Also, the uncertainty bounds between neighboring data points scale up dramatically, leading to an overestimation of the model's uncertainty. 

The shape of the mean and deviation bounds are a result of the Fourier series nature of the underlying circuit kernel, where additional layers introduce higher frequencies into the spectrum of the sum. Therefore, we again stress the importance of finding suitable hyper-parameters and fitting the chosen model to the underlying task, as small differences in the model architecture can lead to large differences in the resulting uncertainty quality as shown in \cref{fig:gp-different-circuit-lengths}.

\section{Classification results} \label{app:class}
\begin{figure}[htbp]
\vskip 0.1in
\begin{center}
\centerline{\includegraphics[width=\columnwidth]{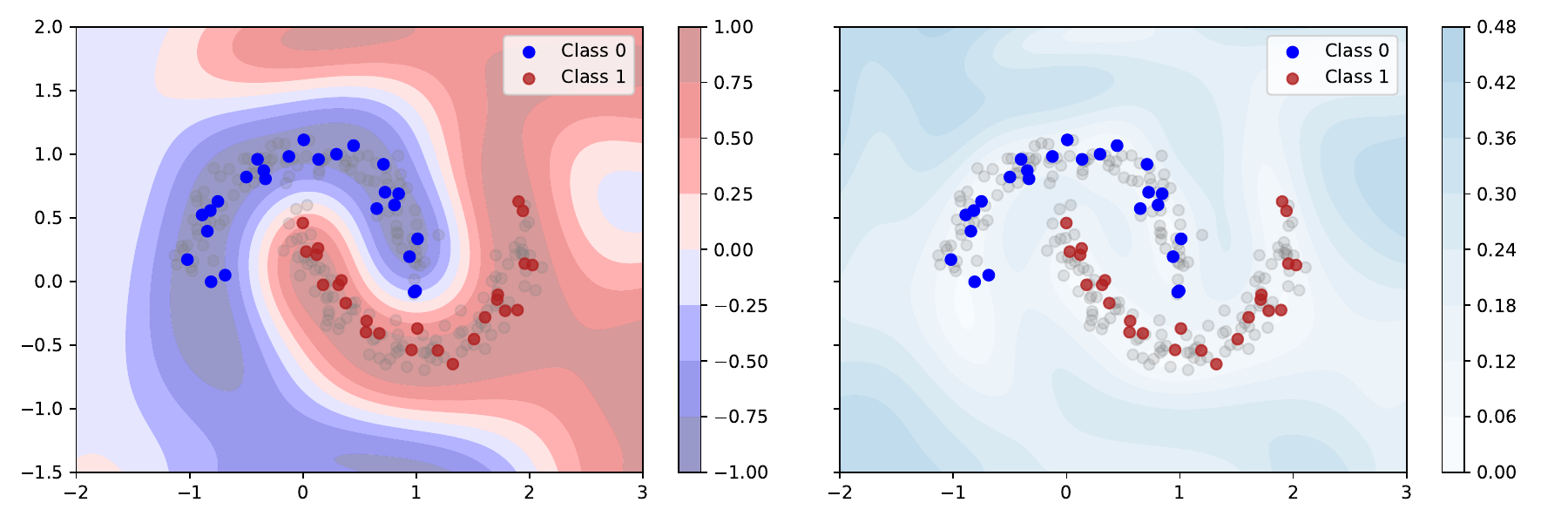}}
\caption{Classification results for a Bayesian \ac{QML} model. Left: Class prediction scores. Right: Predicted model uncertainty (standard deviation) as derived in \cref{subsec:mapping-BayesQC}.}
\label{fig:class-bayes}
\end{center}
\vskip -0.2in
\end{figure}

In the following, we give the results for all mentioned models on the classification experiments. More specifically, for each approach (Bayesian \acs{QML}, [Bias-only, Rotation, Gaussian]- MC dropout, quantum ensemble, quantum \acs{GP}), we first evaluate the predictive performance as signaled by the class scores. Second, we also plot the predictive uncertainty for the output space, as given by the predicted variance. For an example, see the output scores for Bayesian QML in \cref{fig:class-bayes}.

\begin{figure}[htbp]
\vskip 0.1in
\begin{center}
\centerline{\includegraphics[width=\columnwidth]{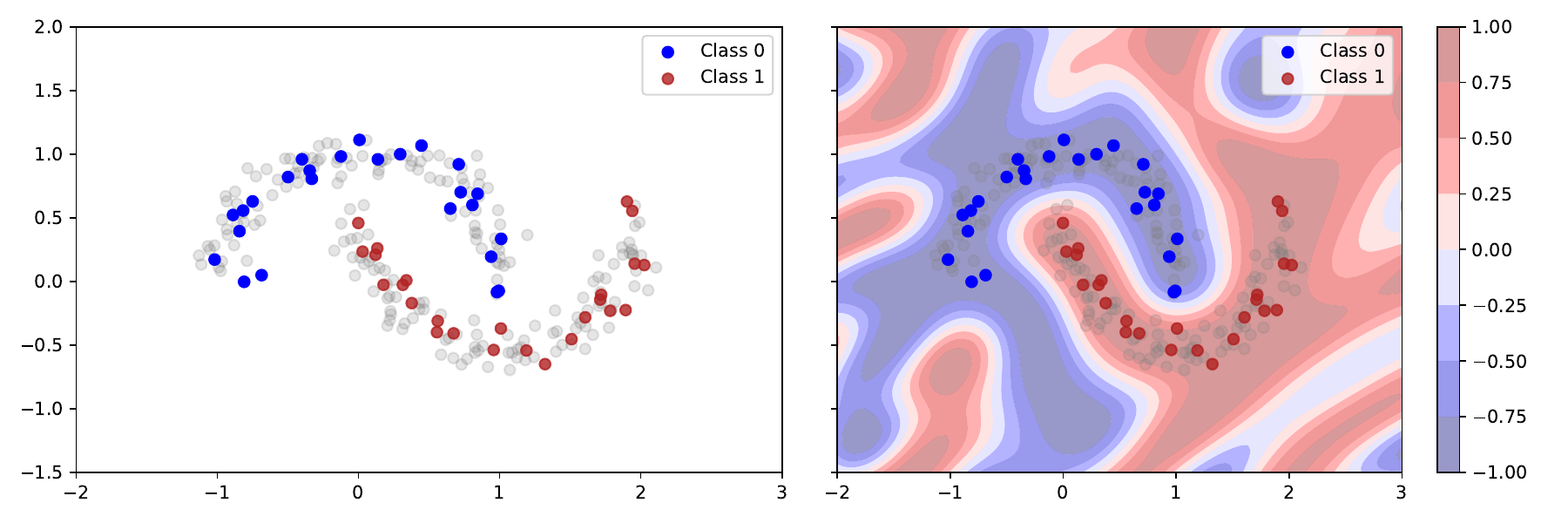}}
\caption{Classification results for deterministic \acs{QML} model using one-qubit six-layer reupload encoding architecture.}
\label{fig:determ-class}
\end{center}
\vskip -0.2in
\end{figure}

Before evaluating the \acs{UQ} capabilities of all models, we show the classification output of a deterministic \acs{QML} model in \cref{fig:determ-class}. As we can see, a periodic output pattern emerges that perfectly fits the training samples but includes many areas of high class-probabilities despite the lack of data in these regions.

Given this behavior as a baseline to improve, we first investigate the classification output of Bayesian \acs{QML} in \cref{fig:class-bayes}. Compared to the outcome of a deterministic \ac{QML} circuit without \ac{UQ} capabilities  we see a much smoother landscape. 
Additionally, from the plot on the right, we can convince ourselves of the basic \ac{UQ} properties of the Bayesian \ac{QML} output, where the standard deviation of the predictive posterior is visualized as blue shades of increasing opacity. In contrast to a simple classification output giving class scores for `0' or `1', the Bayesian treatment of the classifier allows an estimate of uncertainty for each data point, depending on the relative position to the training points and the output function\footnote{Note that the underlying \ac{QML} model is limited to Fourier series functions after all, so we cannot completely eliminate periodicity in the output.}.

\begin{figure}[htb!]
\vskip 0.1in
\begin{center}
\centerline{\includegraphics[width=\columnwidth]{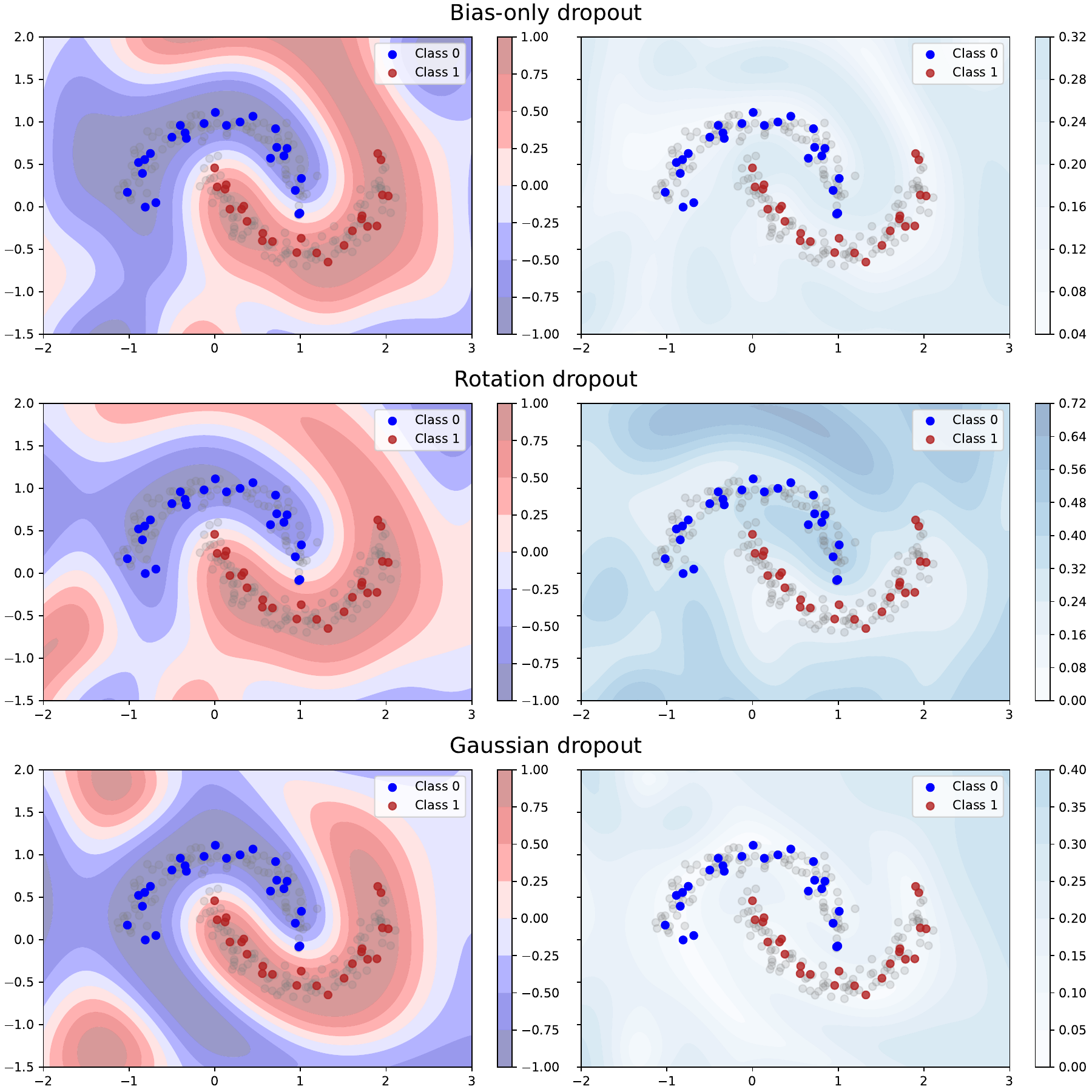}}
\caption{Classification results for three different approaches to \acs{MC} dropout in \acs{QML}.}
\label{fig:class-mc}
\end{center}
\vskip -0.2in
\end{figure}

For quantum \acs{MC} dropout, we visualize the classification output in \cref{fig:class-mc}. First, we can note that the overall output shape of the classification scores is similar between the models (and also with respect to the other models used in the \ac{UQ} experiments), but much smoother than the deterministic baseline from \cref{fig:determ-class}. 
However, we see some differences when investigating in more detail: As in the regression setting, we notice that the uncertainty scale of the models differs (as seen on the right subplots of each row in \cref{fig:class-mc}), where the rotation dropout model has large regions with standard deviations up to $0.7$, whereas the maximum of the other models is around $\sigma=0.35$. A region of interest is the center of the scatter plot, around the position $(x_1,x_2)=(1,0)$, where class `0' (blue) has its right lower tail. In this region, we can see that the rotation dropout has large uncertainty values, even though we have data samples in the training set, a behavior not seen as pronounced in the other two models. Last, regardless of the employed dropout mechanism, we cannot fully eliminate the periodic output of the \ac{QML} classifier, but the dropout averaging scheme helps to smooth out the diverse individual outputs, such that we have only few regions with both a definite class prediction (dark red or blue areas on the left subplot) and a low variance (white region in the right subplot).

\begin{figure}[ht!]
\vskip 0.1in
\begin{center}
\centerline{\includegraphics[width=\columnwidth]{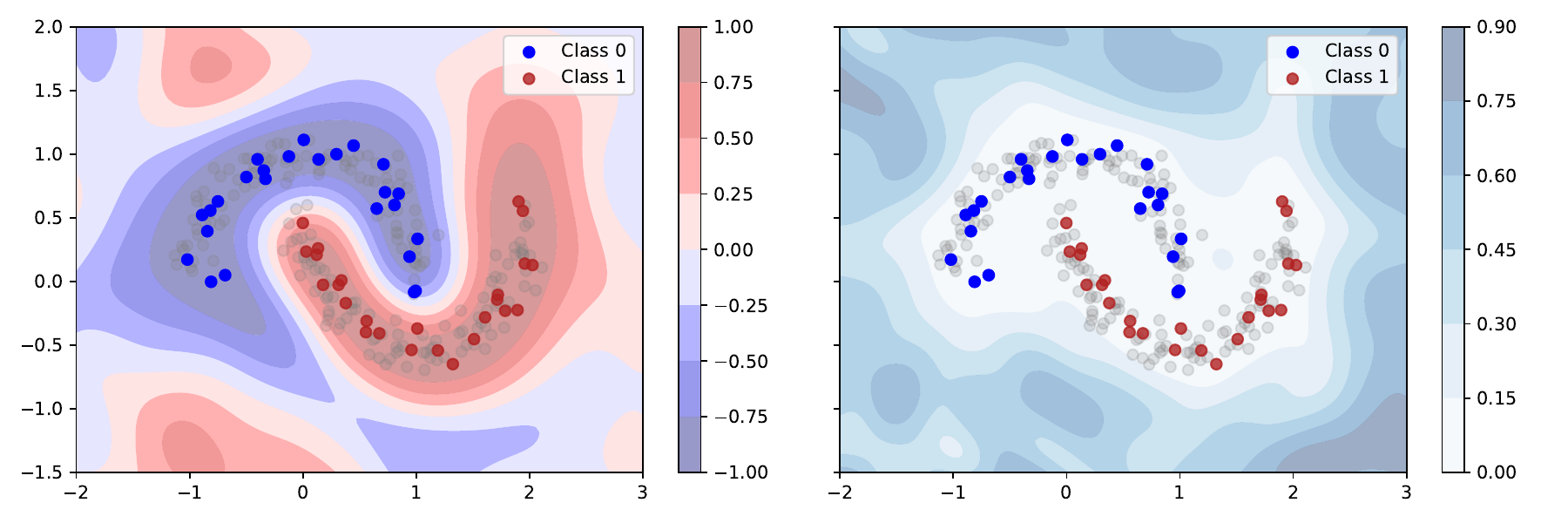}}
\caption{Classification results for an ensemble of eight quantum classifiers using empirical mean and standard deviation to aggregate the outputs.}
\label{fig:class-ensemble}
\end{center}
\vskip -0.2in
\end{figure}
For quantum ensembles (more specifically, the aggregation of eight quantum circuits as used in the regression setting), the classification scores and uncertainty predictions are depicted in \cref{fig:class-ensemble}. While the main class prediction shapes are similar to the methods presented before, two characteristics stand out. On the left plot, we see a comparatively large area shaded in very light tones, indicating an indifferent model output for class predictions far away from observed data points, which already provides some insight into the model's calibration without considering the uncertainty prediction itself. In the right plot, we find that the uncertainty of the model is low in the entire center region, regardless of the class boundaries. This quality presents the difference between ensemble circuits and the other methods discussed here: In regions where many data points are present, the individual model predictions will match almost perfectly, and we obtain very strong confidence values in the model (which also leads to a model calibration that is oriented towards overconfidence -- as seen in \cref{fig:ece-all}). In contrast to circuit ensembles, we do not expect this phenomenon to appear as pronounced, e.g., in \ac{MC} dropout circuits, as each new dropout configuration leads to a new circuit configuration and quantum state resulting in a slightly different output, even in high-density data regions.

\begin{figure}[ht!]
\vskip 0.1in
\begin{center}
\centerline{\includegraphics[width=0.5\columnwidth]{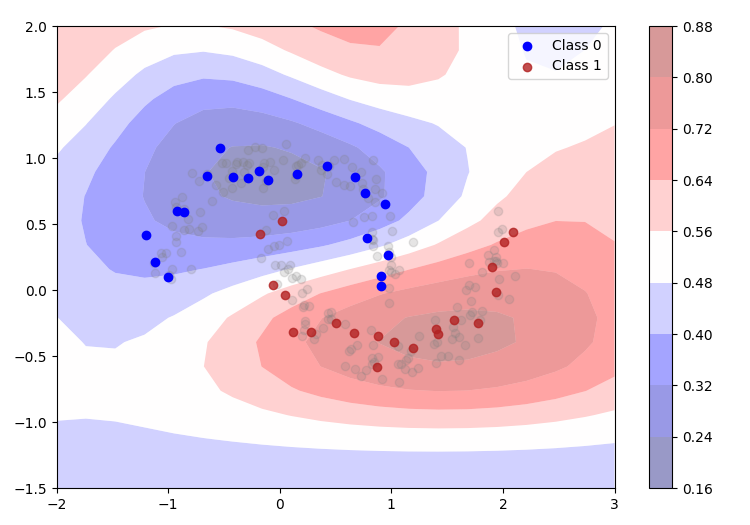}}
\caption{Classification results for a \acl{GP} with a quantum kernel consisting of two qubits and two layers.}
\label{fig:class-gp}
\end{center}
\vskip -0.2in
\end{figure}

Last, we examine the quantum \acs{GP} classification model, using the same two-qubit, two-layer kernel as in the regression case. To this end, we make use of scikit-learn's \texttt{GaussianProcessClassifier}\footnote{\url{https://scikit-learn.org/dev/modules/generated/sklearn.gaussian_process.GaussianProcessClassifier.html}}, which in turn incorporates a logistic link function into the underlying \ac{GP} model, thus mapping the outputs to the range $\hat{y}_k\in [0,1]$. This architecture leads to uncertainty estimates in the class scores but does not allow to explicitly return the uncertainty values as in the approaches before, hence we only show the \acs{GP} class scores in \cref{fig:class-gp}.

As seen in the output landscape, the given quantum kernel is not complex enough to obtain the two-moon shape for a perfect data fit.
While most points are classified correctly and we find areas of large uncertainty (as shown by regions shaded in white), a similar periodic output pattern like the ones discussed in other classification plots emerges.

\section{Limitations and Future Work}
\label{app:limit}

Building upon some of the limitations already mentioned in the main text, we first note the limitation of only integrating a comparatively small subset of the multitude of possible approaches to \ac{UQ} presented in the field of classical \ac{ML}. To give some examples which we find interesting but considered to be out of scope for this work, we mention a list of three possible extensions:
\begin{itemize}
    \item The theory of obtaining uncertainty predictions from deterministic networks gains much attention in classical \ac{ML} and presents itself suitable for integrating into \ac{QML}. Most prominently, Spectral-Normalized Gaussian Processes seek to enforce a bi-Lipschitz condition on the latent model space to obtain distance-awareness \cite{liu2020, liu2022}. Mapping this approach to \ac{QML} with its large latent space can be an interesting topic for future work.
    \item Variational \ac{MC} dropout, an extension of the Gaussian dropout case which incorporates adaptive dropout rates into the model \cite{kingma2015}, may lead to even better results and adaptability as the discussed \ac{MC} dropout methods evaluated here. As we have already seen the nice integration of dropout to \ac{QML}, this may also be an example of a good symbiosis for future insights.    
    \item Lastly, the combination of different approaches, as briefly mentioned when discussing quantum ensembles (and the possibility of integrating probabilistic outputs into each sub-model within the ensemble), is also an interesting topic that might lead to better results for some use cases and underlying data distributions.
\end{itemize}

Another main limitation, also stemming from the fact that we currently only have limited expressive power in the \ac{QML} circuit architectures, is the focus on low-dimensional datasets. While the used data distribution (one-dimensional trigonometric regression and the two-moons classification) allows meaningful visualizations of the model predictions and uncertainty estimates, we must transfer these findings to more advanced, real-world examples before being able to evaluate the model architectures in practice. As a first step, it may be sensible to increase the dimensionality slightly, e.g., using image classification and higher-dimensional regression functions. Parallel to the importance of finding different evaluation datasets, it is essential to vary the underlying \ac{QML} model and scale up the used circuit architecture to more qubits and layers to explore if insights from this work also hold in more complex scenarios.

\section{Code availability}

All implemented model architectures and their respective results on the chosen datasets are made available upon reasonable request.